\documentclass[runningheads]{llncs}
\usepackage{graphicx}
\usepackage{subfigure} 
\usepackage{float}
\usepackage{algorithm}  
\usepackage{algorithmic}  
\begin{document}
\title{Scaffolded Learning of In-place Trotting Gait for a Quadruped Robot with Bayesian Optimization} 
%
%
\author{Keyan Zhai\and
Chu'an Li \and
Andre Rosendo}
\authorrunning{K. Zhai et al.}
%
\institute{School of Information Science and Technology, ShanghaiTech University, \\393 Middle Huaxia Road, Shanghai 201210, China \\
\email{\{zhaiky,licha,arosendo\}@shanghaitech.edu.cn}}
\maketitle              
\begin{abstract}
During learning trials, systems are exposed to different failure conditions which may break robotic parts before a safe behavior is discovered. Humans contour this problem by grounding their learning to a safer structure/control first and gradually increasing its difficulty. This paper presents the impact of a similar supports in the learning of a stable gait on a quadruped robot. Based on the psychological theory of instructional scaffolding, we provide different support settings to our robot, evaluated with strain gauges, and use Bayesian Optimization to conduct a parametric search towards a stable Raibert controller. We perform several experiments to measure the relation between constant supports and gradually reduced supports during gait learning, and our results show that a gradually reduced support is capable of creating a more stable gait than a support at a fixed height. Although gaps between simulation and reality can lead robots to catastrophic failures, our proposed method combines speed and safety when learning a new behavior.

\end{abstract}
\section{Introduction}
Compared to traditional wheeled robots that can only travel on plain grounds, quadruped robots have stronger mobility because of the ability to overcome rough terrains with vertical movement of the legs. However, legged locomotion is much more difficult to model and control for its complexity. Although several controllers for quadruped robots have been proposed~\cite{ref_article1,ref_article2,ref_article3}, the configuration and tuning of their parameters still remain a challenging problem. It is extremely difficult to tune these parameters manually, especially when the robot is operating in the physical world with unpredictable noise. Therefore, parametric search for the gait controllers has become the key to control the locomotion of quadruped robots. 

Various machine learning and optimization methods have been utilized to find these parameters. Such approaches include genetic algorithms~\cite{ref_article4}, Covariance Matrix Adaptation Evolution Strategy (CMA-ES)~\cite{ref_article5}, gradient descent methods~\cite{ref_article6}, and Bayesian Optimization (BO)~\cite{ref_article7,ref_article8,ref_article9}, etc. BO is a sample-efficient black-box global optimization algorithm that is especially suitable when the objective function is expensive to evaluate~\cite{ref_article10}, which is usually the case in the context of robotics. Performing experiments with real robots are usually time-consuming, and the consistency between different iterations is hard to maintain because of the wear and tear of the hardware. Therefore, BO has more advantages over the other methods mentioned above in gait optimization for quadruped robots. 

Most of the previous studies on the gait optimizations using BO focus on the influence of different configurations of the BO algorithm itself. For example, ~\cite{ref_article8} incorporates domain knowledge to reduce dimensionality for BO in higher dimensions of the parameters. ~\cite{ref_article9} proposes an Alternating Bayesian Optimization (ABO) algorithm that iteratively learns the parameters through interactive trials, resulting in sample efficiency and fast convergence. In this work, we focus on the influence of external factors during the optimization process rather than the BO algorithm itself. We studied the impact of supports provided to the robot during the learning process based on the psychological theory of instructional scaffolding.

Instructional scaffolding is a concept in education and learning science, which is used by the teachers to give students supports throughout the learning process. It is widely considered an essential element in effective teaching, and promotes a deeper level of learning than many other common teaching strategies~\cite{ref_article11}. There are three essential features of scaffold learning~\cite{ref_article12,ref_article13}: 1) the interaction between the learner and the expert, 2) learning should take place in the learner's zone of proximal development, and 3) the scaffold provided by the expert is gradually removed. In our work, we also take these three features into account and created similar supporting conditions for the robot. 

In this paper we probe the benefits of scaffolding the learning process of a robot. We assess different support settings to our robot and use Bayesian Optimization to conduct a parametric search towards a stable Raibert controller. In our experiments the robot tries to find suitable controllers to keep itself stable while we gradually decrease the amount of support for this robot. After testing two support conditions we conclude that that a gradually reduced support is capable of creating a more stable gait than a support at a fixed height. Although gaps between simulation and reality can lead robots to catastrophic failures, our proposed method combines speed and safety when learning a new behavior.

\section{Methods}
\label{sec:Methods}

In this work we use a robot, called Slimdog, to test the influences of supports on the learning of a stable, Raibert-based controller. The robot consists of a combination of carbon fiber rods and 3d printed ABS parts, and has a total weight of 2.1 kg. The total degrees of freedom of Slimdog is 12 with 3 per leg. The actuation of the robot is done with 12 over-the-counter servo motors with an output torque of 35 kg.cm, and the entire system is powered by a Lipo battery located at the center of the robot. The electronics of the system are controlled by an Arduino board, which sends PWM signals to the servo motors and reads signals from the IMU and the strain gauge located at the support. The robot communicates with a computer through a serial port, which is running Bayesian Optimization to decide the next set of parameters to be evaluated. The dimensions of Slimdog are 48.5 cm long, 42 cm wide and 63 cm with its legs fully extended.

\subsection{Quadruped Gait Controller}

The controller of the robot is based on the model proposed in~\cite{ref_article3} by Marc Raibert. There are three key aspects of the controller: 1) the hopping height of the leg, 2) the body attitude during stance, and 3) the forward running speed. 

In our case, we only tested the motion of the robot running in place, so we mainly considered the first two aspects of the model, and encoded the parameters in the following way:

\begin{table}
\centering
\caption{Parameter settings of the gait controller}\label{tab1}
\begin{tabular}{|l|l|l|}
\hline
Meaning &  Parameters & Range\\
\hline
Hopping height $h$ (mm) &  $x0$ & $[60,120]$\\
\hline
gain $k_{p1}$ for controlling pitch &  $x1$ & $[-1,1]$\\
\hline
gain $k_{v1}$ for controlling pitch & $x2$ & $[-1,1]$ \\
\hline
gain $k_{p2}$ for controlling roll & $x3$ & $[-1,1]$\\
\hline
gain $k_{v2}$ for controlling roll & $x4$ & $[-1,1]$\\
\hline
\end{tabular}
\end{table}

During the stance phase, the legs on the ground rotate $\theta_{pitch}$ and $\theta_{roll}$ to adjust the body attitude of roll and pitch, which can be calculated by the following formulas:  

$$\theta_{pitch} = - k_{p1}(\phi_P-\phi_{Pd})-k_{v1}(\dot{\phi_{y}})$$
$$\theta_{roll} = - k_{p2}(\phi_R-\phi_{Rd})-k_{v2}(\dot{\phi_{x}})$$

where $\phi_P$ stands for the pitch angle, $\phi_{Pd}$ stands for the desired pitch angle, and $\dot{\phi_{y}}$ stands for the angular speed around the $y$ axis. Similarly, $\phi_R$ stands for the roll angle, $\phi_{Rd}$ stands for the desired roll angle, and $\dot{\phi_{x}}$ stands for the angular speed around the $x$ axis. All of these data can be read from the IMU equipped on the robot. $\phi_{Pd}$ and $\phi_{Rd}$ are set to 0 for the balance of the robot.

\subsection{Bayesian Optimization}


Bayesian Optimization uses Gaussian Process to create a probabilistic surrogate model of the unknown objective function. Gaussian Process is a collection of random variables, any finite number of which have (consistent) joint Gaussian distributions~\cite{ref_article14}. It can be defined by its mean function $m(x)$ and covariance function $k(x, x')$, and represented as:

$$f \sim \mathcal{G}\mathcal{P} (m,k)$$

In each iteration, BO samples a point selected by the acquisition function, and update the surrogate model created by GP until certain number of iterations is reached. We used Upper Confidence Bound (UCB) as the acquisition function of BO, and implemented the algorithm with Python. The objective function in our experiments is set to be the percentage of weight supported by the robot itself.

\subsection{Experimental Setup}


In our experiments we connected a strain gauge to a platform above the robot and used two ropes to connect the robot to this strain gauge. The ropes hang the robot from its bottom and were used to prevent the robot from collapsing to the floor when the controller failed to coordinate the legs and keep the robot upright, while the strain gauge is essential to measure the amount of force sustained by the support system. We then created two experimental conditions:

\begin{itemize}
\item[*] Minimum support - The robot started at 39 cm above the ground in the first 50 iterations, and in the last 10 iterations we lowered the support to 32.5 cm above ground (condition at which the rubber only supports the system when the robot reaches a very unstable gait).
\item[*] Reducing support - The height of the support structure starts at 47.5 cm and is maintained at this height for 10 iterations. In this height the robot has a stronger tendency to extend the knee joint to reduce the need for support. The height of the support is reduced to 45 cm from 11 to 20, 42 cm from 21 to 30, 39 cm from 31 to 50, and the last 10 iterations are at 32.5 cm from the floor.
\end{itemize}

\begin{figure}[th!]
\centering
\subfigure[0 ms]{
\includegraphics[scale=0.25]{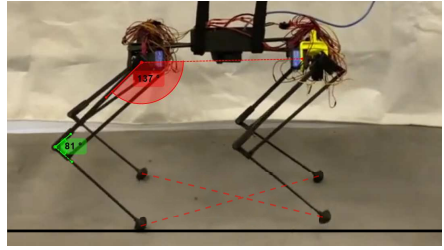} \label{1}
}
\quad
\subfigure[150 ms]{
\includegraphics[scale=0.25]{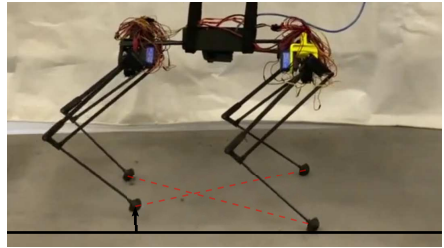} \label{2} 
}
\quad
\subfigure[300 ms]{
\includegraphics[scale=0.25]{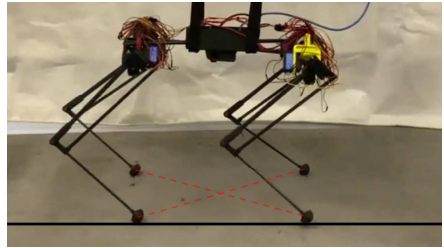}\label{3}
}
\caption{Snapshots of the experiments. Following a variant of Raibert's control the robot uses the virtual leg principle to control opposing legs to emulate a virtual leg touching the floor. The control parameters are chosen according to the Bayesian Optimization algorithm. }
\label{fig1}
\end{figure}

The force value registered by the strain gauge was used to estimate the fitness of the system (we adopted the sum of the inverse of this force registered over the experiment), and video recordings from the experiment were later used in combination with Kinovea to calculate the vertical translation of the center of gravity and the changes in joint angle during the experiment. The fitness in our experiments stands for the percentage of weight supported by the robot itself rather than by the ropes. Higher fitness value means more support from the robot, which can be viewed as a better performance. A demonstration of our experimental setting can be found at Fig.~\ref{fig1}.

\section{Results and Discussion}

\begin{figure}[ht!]
\includegraphics[width=\textwidth]{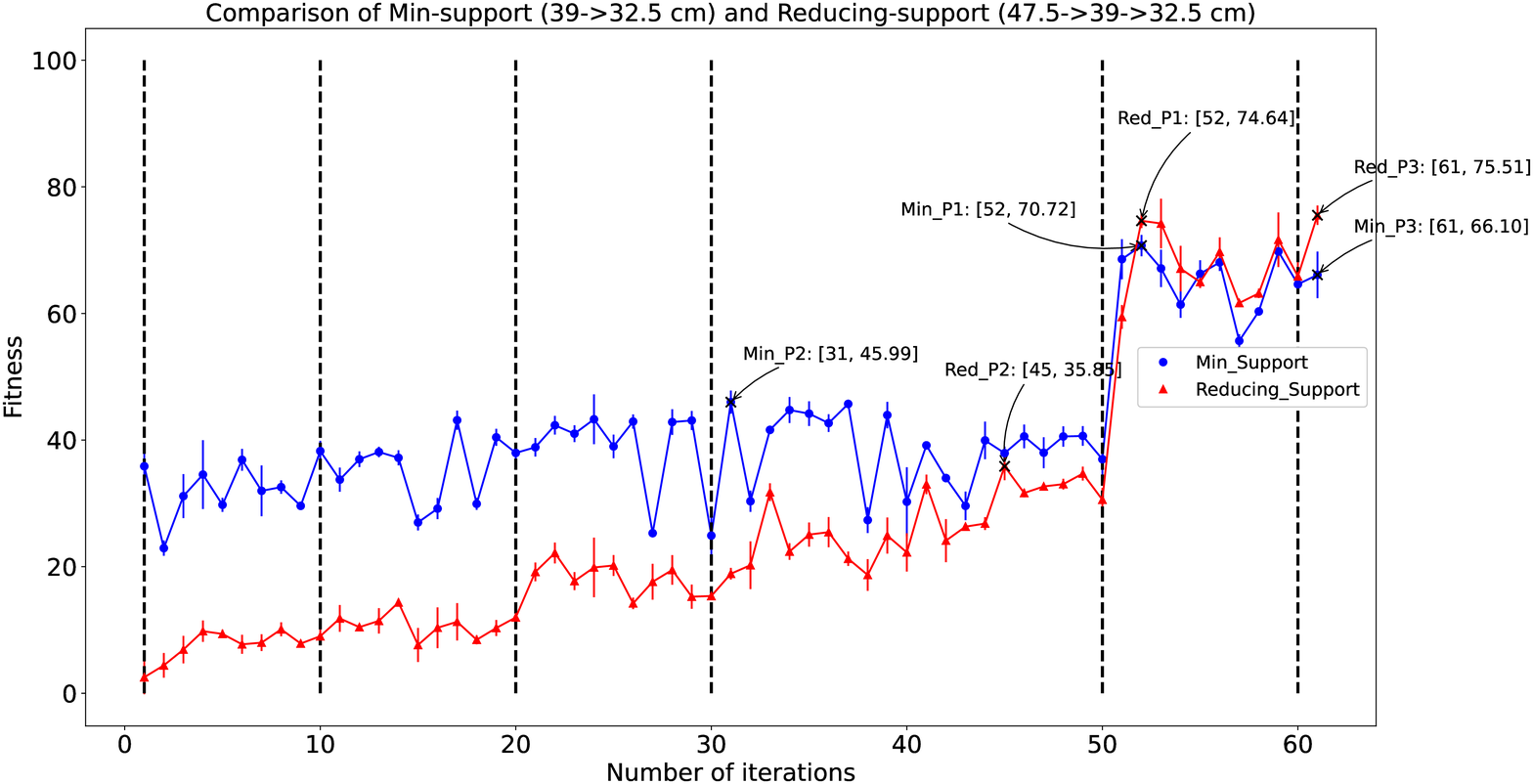}
\caption{Comparison between the fitness provided with minimum support and reducing support. In the figure, ``Min/Red\_P1" represents the best fitness through iteration 1 to 60, ``Min/Red\_P2" represents the best fitness through iteration 1 to 50, and ``Min/Red\_P3" represents the fitness using the same parameters as ``Min/Red\_P2" (best parameters in the first 50 iterations), but with the height of robot to ground as 32.5 cm.} \label{fig2}
\end{figure}

The whole optimization process of the experiments is shown in Fig.~\ref{fig2}. From the results, we see that for the first 50 iterations, the fitness learned by the minimum support is generally higher than that learned by the reducing support. An underlying reason for this difference is related to the weight of the robot being supported by the ropes instead of the robot itself. As iteration proceeds, the support from the ropes reduces, and the robot relies more and more on itself, resulting in a rising tendency of the fitness. During the last 10 iterations, where a sudden spike in values happens due to the lower contribution of the support in bearing the weight, the fitness learned by the reducing support generally surpasses the fitness learned by the minimum support. As shown in the figure, the fitness of the best gait learned by reducing support (represented as ``Red\_P1") is slightly higher than that learned by the minimum support (represented as ``Min\_P1"). The learned control parameters for the optimal fitness ``Red\_P1" and ``Min\_P1" are the same values, and are shown in Table~\ref{tab2}. The reason why the same control parameters lead to different fitness results might be due to random errors and noises. Also, the reducing support provides more protection to the robot during the learning process, and there is less wear and tear of the robot.


\begin{table}
\centering
\caption{Parameter for the optimal fitness ``Red\_P1" and `Min\_P1"}\label{tab2}
\begin{tabular}{|l|l|}
\hline
Parameters & Value\\
\hline
$x0$ & $77.99391797172291$\\
\hline
$x1$ & $-0.08307456005155922$\\
\hline
$x2$ & $0.17227550937265934$ \\
\hline
$x3$ & $0.4861753588049966$\\
\hline
$x4$ & $-0.05160482932681609$\\
\hline
\end{tabular}
\end{table}

\begin{figure}[ht!]
\includegraphics[width=\textwidth]{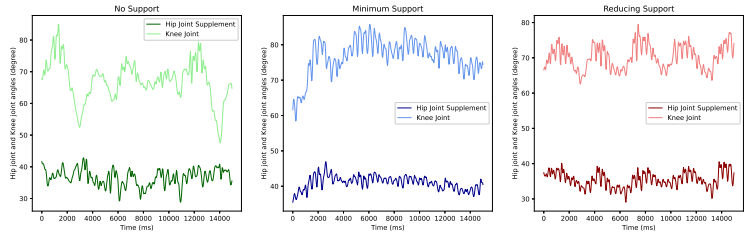}
\caption{The comparison of the hip and knee joint angles between ``No\_P1", ``Min\_P1", and ``Red\_P1". For the hip joints, supplement angles are used for clarity.} 
\label{fig3}
\end{figure}

We also performed experiments of the robot starting directly from 32.5 cm with no prior iterations, represented as ``No\_P1" and the green color. Fig.~\ref{fig3} shows the hip and knee joint angles of ``No\_P1", ``Min\_P1", and ``Red\_P1". During the 15-second run, the joint angles learned by reducing support show a cyclical pattern with a period of roughly 3 seconds. The joint angles learned by no support and minimum support do not show a repeating pattern, but the joint angles learned by no support vary more than the other two, which shows a more irregular gait. It is quite difficult to replicate a condition that can be truly called without support, as we observed in undocumented experiments that our learning experiments without support do not last more than 2 seconds. We noticed that even with the lowest adopted support, of 32.5 cm above the ground, the robot still finds ways to tug against the structure when at the brink of collapse.

\begin{figure}[ht!]
\includegraphics[width=\textwidth]{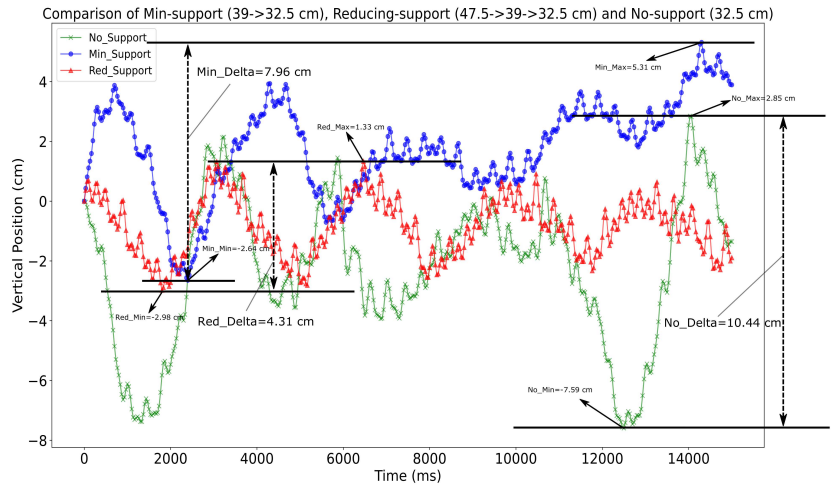}
\caption{The comparison of vertical positions of the center of gravity between ``No\_P1", ``Min\_P1", and ``Red\_P1". } 
\label{fig4}
\end{figure}

Fig.~\ref{fig4} shows the vertical positions of the center of gravity of the robot. There is also a cyclical pattern of ``Red\_P1" with a period of roughly 3 seconds, which ``No\_P1" and ``Min\_P1" do not have. The difference between maximum and minimum vertical positions are represented as ``No/Min/Red\_Delta", and can tell the overall stability of the gaits. We can see that the gait learned by the reducing support has more stability than those learned by no support and minimum support.

\section{Conclusion}

In this work we developed a lightweight quadruped robot, implemented a Raibert controller with a few variable parameters to be decided by a Bayesian Optimization algorithm, and explored parametric possibilities for those while constraining our robot to a scaffold. Although we notice that learning without a scaffold lead our robot to catastrophic failures, we adopted two scaffolded methods to safely conduct the parametric training that would converge to a higher stability.

We found that a gradually decreasing support system is capable of reaching better results than a system with a constant support, and both systems are superior to the absence of a support as they ensure the integrity of our robot until the end of the heuristic training. While simulations are a powerful tool to explore controllers before deploying those to real robots, roboticists should be aware that those will not work in the same manner that they performed inside simulations, and our proposed method combines a model-free search (Bayesian Optimization) to a model-based controller (Raibert Controller) for a safe and fast real world implementation.

%
%
%
%

\end{document}